\documentclass[11pt]{article}
\usepackage{geometry}
\usepackage{coling2020}
\usepackage{times}
\usepackage{url}
\usepackage{latexsym}
\usepackage{microtype}
\usepackage[hidelinks]{hyperref}
\usepackage{array}
\usepackage{graphicx}
\usepackage{listings}
\usepackage{float}
\usepackage{caption}
\usepackage{url}
\usepackage{color}
\usepackage[table,x11names]{xcolor}

\colingfinalcopy 

\newcommand\firstApp{BiLSTM + Attention}
\newcommand\secondApp{Transformers}

\usepackage{soul}
\usepackage[caption=false]{subfig}

\newcommand\nnfootnote[1]{%
  \begin{NoHyper}
  \renewcommand\thefootnote{}\footnote{#1}%
  \addtocounter{footnote}{-1}%
  \end{NoHyper}
}

\title{IITK at SemEval-2020 Task 10:  Transformers for Emphasis Selection}

\author{Vipul Singhal$^{*}$ \qquad    
  Sahil Dhull$^{*}$  \qquad  
  Rishabh Agarwal\thanks{\quad The authors equally contributed  to this work.} \qquad
  Ashutosh Modi  \\
{Indian Institute of Technology Kanpur (IITK)} \\
  {\tt \{vipulsn,sahild,rish\}@iitk.ac.in}  \\
  {\tt ashutoshm@cse.iitk.ac.in}  \\
}

\date{}

\begin{document}
\nnfootnote{\noindent This work is licensed under a Creative Commons Attribution 4.0 International Licence. Licence details: \\http://creativecommons.org/licenses/by/4.0/.}
\maketitle
\begin{abstract}
This paper describes the system proposed for addressing the research problem posed in Task 10 of SemEval-2020\footnote{ \href{https://competitions.codalab.org/competitions/20815}{https://competitions.codalab.org/competitions/20815}}: Emphasis Selection For Written Text in Visual Media. We propose an end-to-end model that takes as input the text and corresponding to each word gives the probability of the word to be emphasized. Our results show that transformer-based models are particularly effective in this task. We achieved the best Match$_{m}$ score (described in section \ref{sec:metric}) of 0.810 and were ranked third on the leader-board.
\end{abstract}

\section{Introduction}
\label{intro}
\begin{figure}%
    \centering
    \subfloat[Less Impact]{{\includegraphics[width=4cm, height = 5.5cm]{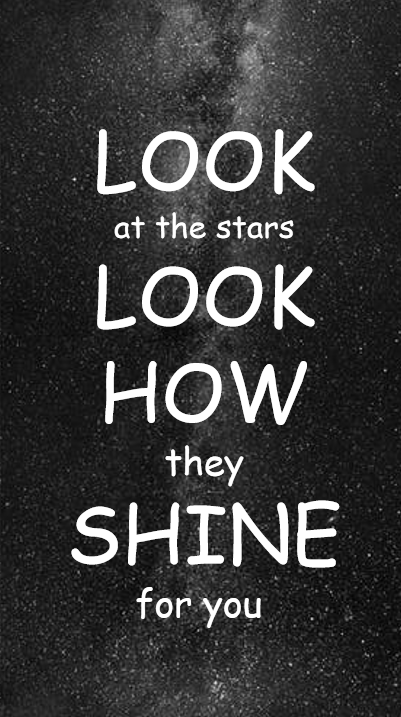} }}%
    \qquad
    \subfloat[More Impact]{{\includegraphics[width=4cm, height = 5.5cm]{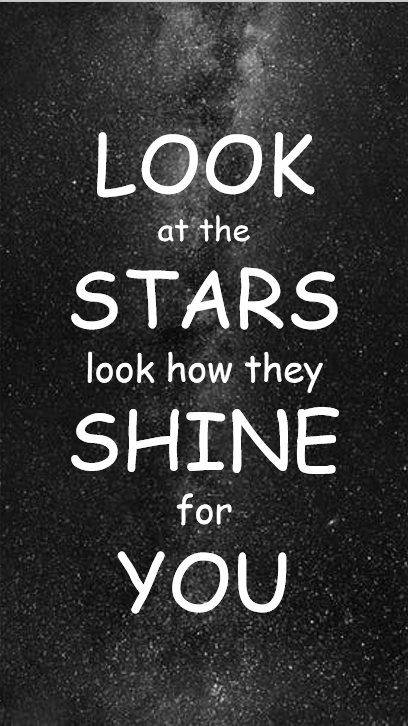} }}%
    \caption{Emphasizing different sets of words}%
    \label{fig:comparison}%
\end{figure}

Visual communication relies on images and short texts, e.g., in flyers, posters, etc. The purpose of these is to convey the message effectively and without ambiguity. Moreover, they should be able to attract the reader's attention within the first few seconds. For text, this can be achieved by laying emphasis on particular words to convey the intent better. The emphasis selection task is about designing automatic methods for choosing candidate words to be emphasized in short written texts, to enable automated design assistance in authoring (Figure \ref{fig:comparison}). 
The dataset provided for this task is in English, and task description paper \cite{shirani2020semeval} provided by the organizers describes the  task, data, evaluation, results, and a summary of participating systems.

We tried two different approaches for the task. Our \firstApp{} approach is inspired by the baseline paper \cite{main}. In this approach, we tweaked the BiLSTM layers \cite{lstm}, tried other layers like GRU \cite{gru}, and used character embeddings along with word embeddings \cite{char_emb}.

Our \secondApp{} approach involves transfer learning using Transformer based models \cite{transformer}. This approach involves two types of models, the first one being a transformer-based model with the BiLSTM layer, the attention layer \cite{attention}, and fully connected layers on top. The second type of model involves transformer-based models with fully connected layers. We used BERT \cite{bert}, RoBERTa \cite{roberta}, XLNet \cite{xlnet}, and GPT-2 \cite{gpt2} as transformer-based models. In the end, we tried the homogeneous and heterogeneous ensemble of these models.

Due to the nature of the problem, the systems which incorporate the whole context of a sentence would perform better. Further, the small size of the dataset was a bottleneck that can be countered by using transfer learning via the pre-trained models. Using transformer-based models accounts for both of these observations.

Our best submission achieved Match$_{m}$ score of 0.810 (Match$_{m}$ evaluation metric is described in Section 2.2) and was ranked third on the leaderboard. Our submissions under-performed in Score 1 (Match\(_{1}\) as defined in Section \ref{sec:metric})
as compared to the other top-performing teams. Our code is available on Github\footnote{\ \href{https://github.com/SahilDhull/emphasis_selection}{github.com/SahilDhull/emphasis\_selection}}.

\section{Background}

\subsection{Problem definition}

Given a sequence of words or tokens C = $\{ x_1, x_2, ..., x_n \}$ in a text, the task is to compute a probabilistic score $S_i$ for each $x_i$ which indicates the degree of emphasis to be laid on the word.

\subsection{Evaluation Metric}\label{sec:metric}

The evaluation metric for our problem is defined as follows:

For a given m (from 1 to 4), we first define 2 sets, $S_m^{(x)}$ - set of $m$ words with top $m$ probabilities according to ground truth
and $\hat S_m^{(x)}$ - set of $m$ words with top $m$ probabilities according to model predictions. To get $S_m^{(x)}$, each word in the sentence has been manually annotated by 9 annotators using Amazon Mechanical Turk. More details regarding the same can be found in the baseline paper \cite{main}.\\
Based on these 2 sets, we define Match$_{m}$ as
\[ {Match_m = \frac{\sum_{x \in D_{test}} \mid S_m^{(x)} \cap \hat S_m^{(x)} \mid / min(m,\mid x \mid )  }{\mid D_{test} \mid}} \]
where $D_{test}$ is the dataset and $x$ is text instance. We find Match$_{m}$ for m $\in \{ 1,..,4\}$ and finally averaged to obtain the final score.

\subsection{Data}

We used the officially released dataset\footnote{\href{https://github.com/RiTUAL-UH/SemEval2020_Task10_Emphasis_Selection}{https://github.com/RiTUAL-UH/SemEval2020
\_Task10\_Emphasis\_Selection}}, which is the combination of the following two datasets:

\textbf{Spark dataset:} This dataset is a collection of short texts containing a variety of subjects featured in advertisements, posters, flyers, or motivational memes collected from Adobe Spark and contains 1200 instances.

\textbf{Quotes dataset:} This dataset collected from Wisdom Quotes contains 2,718 instances of quotes from well-known authors.

The dataset contains very short texts, usually fewer than 10 words, and is randomly divided into training (70\%), development (10\%) and test (20\%) sets by the organizers.

\subsection{Previous Work}
The baseline paper \cite{main} employs an end-to-end label distribution learning (LDL) and predicts a selection distribution. The model consists of an embedding layer, which is GloVe \cite{glove} or ELMo \cite{ELMo} embeddings, followed by BiLSTM layers and, in the end, fully connected layers. Here, the output of BiLSTM layers for each word is passed through fully connected layers to predict the probabilities of emphasis and non-emphasis whose sum is 1.

\section{System Overview}
Our system takes as input the words in the text and corresponding to each word, gives the probability of the word to be emphasized.
We tried two different types of sequence labeling approaches to learn emphasis patterns. 

\begin{figure}[H]
    \centering
    \subfloat[\firstApp{} Approach Model \label{fig:model1}]{{\includegraphics[width = 10.5cm, height = 5cm]{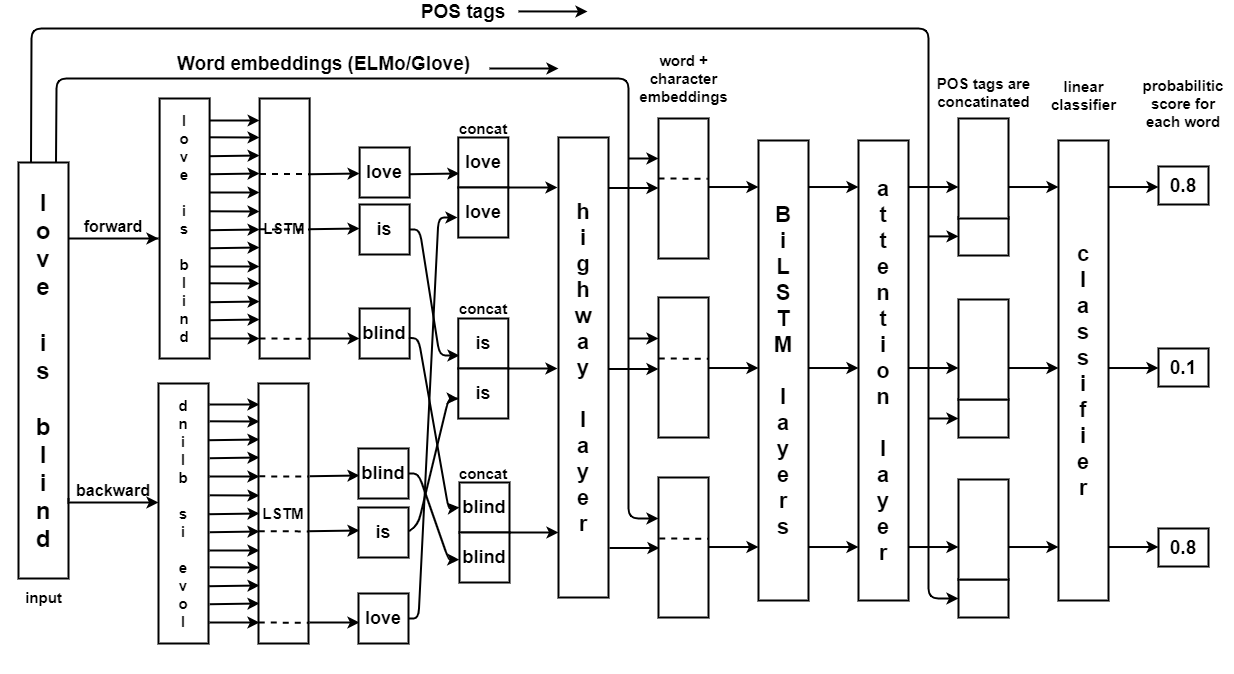}}}%
    \qquad
    \subfloat[\secondApp{} Approach Model \label{fig:bert_classifier}]{{\includegraphics[width = 4.5cm, height = 5cm]{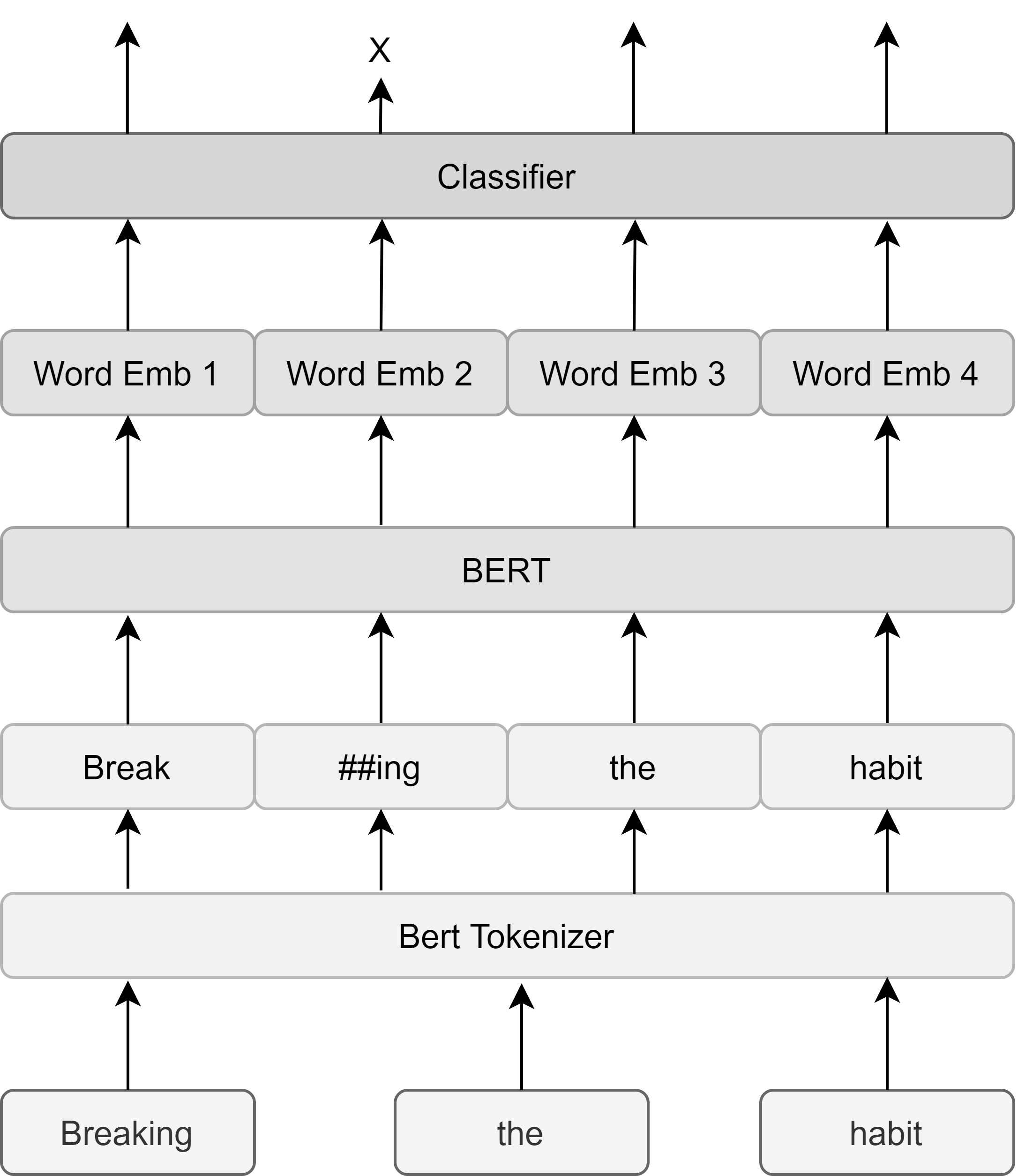}}}%
    \caption{Models for both approaches}%
\end{figure}

\subsection{BiLSTM + Attention Approach}

This approach involves character-level embeddings of each word of a sentence in addition to the word embeddings (Figure \ref{fig:model1}).
The characters of the sentences are passed through a pair of forward and backward LSTM. For each word, the outputs of the forward and backward LSTMs at the position of the last character of the word are taken, concatenated, and then passed through a highway layer to obtain the character-level embeddings of that word. These character-level embeddings are then concatenated with word embeddings obtained using pre-trained models such as GloVe or ELMo and passed through a pair of BiLSTM Layers. A self-attention layer \cite{self_attention} which helps in learning the dependencies between the words in the sentence and gives different importance to different words while predicting
is finally added, followed by a neural-network-based classifier which gives the probability of emphasis for each word. We also concatenated the POS tag of the words to the output of the Attention layers. The representation for each word at the output of Attention layers (along with POS tag concatenation) is passed through the fully connected layers to output the emphasis probability of the word.
\subsection{Transformers Approach}
In this approach, we use Transformer-based models with fully connected layers. Words are tokenized using appropriate tokenizer for each transformer model. Word embedding is obtained by concatenating the hidden layers of all encoder layers 
of the transformer-based model. And in the end, there are few fully connected layers with a dropout layer after each fully connected layer except the last one. Finally, the output of the last layer is passed through a sigmoid layer, which gives the probabilistic score of each word (Figure \ref{fig:bert_classifier}).
Here transformer-based models include BERT and BERT Large, RoBERTa and RoBERTa Large, XLNet Large, GPT-2 Medium, ALBERT, etc. We call this type of model as Transfomer-model and Classifier model (like BERT and Classifier model in Figure \ref{fig:bert_classifier}) in all further discussions.

Finally, we used an ensemble of the above transformer models by taking the average of the scores predicted by these models. This helps to combine multiple models into one predictive model with less variance and improved predictions.

\section{Experimental Setup}

Our implementation uses PyTorch\footnote{\href{https://pytorch.org/}{https://pytorch.org/}} library for deep learning models and the Transformers library by Hugging face\footnote{\href{https://huggingface.co/transformers/}{https://huggingface.co/transformers/}} for the pre-trained transformer models and their tokenizers.

In \firstApp{} approach, for obtaining character-level embeddings, a pair of BiLSTM layers with hidden size of 300 is used, whose output is passed through a highway layer and then concatenated with the GloVe or ELMo embeddings. This concatenated vector is then passed through another pair of BiLSTM layers with hidden dimension of 512. For the classifier, a pair of fully connected layer is used with ReLU activation function \cite{relu} and hidden dimension of 20. Finally, the output of the last fully connected layer is passed through a sigmoid layer, which gives the probabilistic score of each word. To avoid overfitting, dropout layers with a probability of 0.3 were used.

In the \secondApp{} approach, the pre-trained transformer models were used without freezing the layers, and the outputs of all the layers were concatenated. For the classifier, three fully connected layers are used with ReLU activation function and hidden dimension of 900 and 40 for larger transformer models and 300 and 20 for normal models.
Dropout layers with a probability of 0.3 are also added to avoid overfitting.

In both approaches, Binary Cross-Entropy loss is used for training the model, whereas Match\(_{m}\) score is used as the performance metrics for validation (as described in Section 2.2). We used Adam optimizer \cite{adam} with the learning rate set to 0.001 for \firstApp{} approach and 2e-5 for the \secondApp{} approach. The model is fine-tuned for 100 epochs in \firstApp{} approach and 30 epochs in the \secondApp{} approach, and the reported test result corresponds to the best score obtained on the validation set.

\section{Results}
We attempted numerous small changes to our models in the \firstApp{} as well as the \secondApp{} approach to enhance the performance. This included hyper-parameter tuning like changing size and dimensions for different layers with some specific attempts particular to each approach.

\begin{table}[H]
\parbox{.4\linewidth}{
\centering
\begin{tabular}{|m{3.6cm}|c|}
\hline
    \textbf{Model} & \textbf{Best Score} \\
    \hline
    Baseline Model & 0.731\\
    \hline
    Baseline + Character Embedding Model & 0.743\\
    \hline
    Baseline + Character Embedding Model + POS Tag Concatenation & 0.747\\
    \hline
    Baseline + Character Embedding Model + POS Tag Concatenation + Language Model & 0.738\\
    \hline
\end{tabular}
\caption{\label{first_results_table} \firstApp{} Approach}
}
\hfill
\parbox{.59\linewidth}{
\centering
\begin{tabular}{|m{6.3cm}|c|}
\hline
    \textbf{Model} & \textbf{Best Score} \\
    \hline
    BERT + BiLSTM + Attention + FC layers & 0.771 \\
    \hline
    BERT + GRU + Attention + FC layers & 0.755 \\
    \hline
    BERT + Classifier & 0.775 \\
    \hline
    BERT\_Large + Classifier & 0.789 \\
    \hline
    RoBERTa\_Base + Classifier & 0.775 \\
    \hline
    RoBERTa\_Large + Classifier & 0.790 \\
    \hline
    XLNet\_Large + Classifier & \textbf{0.804} \\
    \hline
    ALBERT + Classifier & 0.755 \\
    \hline
    GPT-2 + Classifier & 0.725 \\
    \hline
    XLM-RoBERTa + Classifier & 0.785 \\
    \hline
\end{tabular}
\caption{\label{second_results_table}\secondApp{} Approach}
}
\end{table}

For the \firstApp{} approach, we tried training the model separately on Quotes and Spark dataset. We experimented using highway layers along with BiLSTM layers to attain a better outcome. We also tried using GRU layers \cite{paper2} instead of BiLSTM layers. The word embeddings used in this approach are GloVe and ELMo embeddings. We also tried incorporating a language model to the character embedding model as done in this paper \cite{language_model}, where they predict the next word using forward and backward character RNN layers whenever they encounter space in the sentence. In Table \ref{first_results_table}, we present the best results (evaluation score as defined in section 2.2) on the validation set obtained after all these attempts for the \firstApp{} approach. These were achieved using ELMo embeddings and BiLSTM layers.

For the \secondApp{} approach, we altered the number of freezed layers in all the transformer models (BERT, XLNet, RoBERTa) while training. We attempted taking first-word embeddings/average of embeddings of all words in cases in which a word is broken into multiple tokens by the transformer model tokenizer. We also tried taking the output of the last hidden layer as well as concatenating all hidden layers of the transformer models as the final embedding of the word. Another type of model tried was the BERT Language Model with BiLSTM layers, attention, and fully connected layers on the top. This is basically using all the BERT hidden layers as embeddings by concatenating them, rather than GloVe or ELMo embeddings.

Best results were obtained by concatenating all layers, taking first token output as word embedding, without freezing any layer of the transformer model, and taking three fully connected layers on the top (their dimensions fine-tuned for each model). Table \ref{second_results_table} contains the best results for all models on the validation set in the \secondApp{} approach according to the evaluation metric.

\begin{table}[H]
\begin{center}
\begin{tabular}{|m{6cm}|c|c|}
    \hline
    \textbf{Ensembling Model (runs)} & \textbf{Validation Score} & \textbf{Test score}\\
    \hline
    XLNet (11) & \textbf{0.811} & 0.807 \\
    \hline
    BERT (2), RoBERTa (3), XLNet (8) & 0.808 & 0.809\\
    \hline
    BERT (2), RoBERTa (3), XLNet (9) & 0.810 & \textbf{0.810} \\
    \hline
\end{tabular}\\
\end{center}
\caption{\label{ensembling_results_table} Ensembling Results}
\end{table}

In the end, we tried an ensemble of models from the \secondApp{} approach. We picked BERT\_Large, RoBERTa\_Large, and XLNet\_Large models for both homogeneous and heterogeneous ensembling. After running multiple runs of each model, we took an average of the scores across multiple runs to obtain the final score for each word in each instance. Some of the results on the validation and test set are given in Table \ref{ensembling_results_table} (Large variants of all transformer models were used). The numbers in the bracket denote the number of runs of that particular model that were used in the ensemble.

We ranked third on the task leader board with a test score of 0.810. The top-performing team has a score of 0.823, while the second team has a score of 0.814.

\begin{figure}[H]
    \centering
    \includegraphics[width = 16cm, height = 2cm]{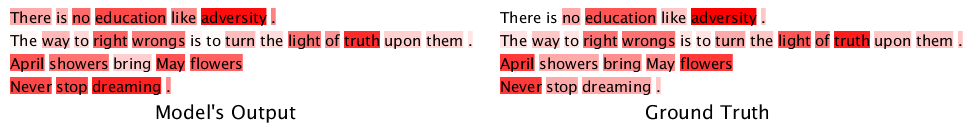}
    \caption{Heatmap of emphases}
    \label{fig:Error_Analysis}
\end{figure}
Figure \ref{fig:Error_Analysis} shows some examples, with a heatmap showing the XLNet\_Large + Classifier model’s predicted score and ground truth probabilities. The model performs well for the first two examples, where the prediction is fairly accurate for all the words. For the last two examples, the model fails for some of the words. In general, the model ceases to perform well on short sentences with three to four words.

\section{Conclusion}
We described the systems used for submission in the Task of Emphasis Selection For Written Text in Visual Media. The task was similar to the Sequence Labeling task, and hence, similar approaches can be used for this task.
Our main approach used Transformer-based models like BERT, RoBERTa, XLNet, and their ensembles. These models are pre-trained and hence, perform better after fine-tuning on our small dataset.

Future work includes making an application that automates the process of poster or advertisement making. A short written text will be the input to the app. It then predicts the probability of various words to be emphasized based on predictions from our model and also adds an appropriate image as the background either using GANs \cite{gan} or using some other API. Finally, it will output a poster, flyer, or advertisement.

\bibliographystyle{coling}
\bibliography{semeval2020}

\begin{thebibliography}{}

\bibitem[\protect\citename{Agarap}2018]{relu}
Abien~Fred Agarap.
\newblock 2018.
\newblock Deep learning using rectified linear units (relu).
\newblock {\em arXiv preprint arXiv:1803.08375}.

\bibitem[\protect\citename{Bahdanau \bgroup et al.\egroup }2014]{attention}
Dzmitry Bahdanau, Kyunghyun Cho, and Yoshua Bengio.
\newblock 2014.
\newblock Neural machine translation by jointly learning to align and
  translate.
\newblock {\em arXiv preprint arXiv:1409.0473}.

\bibitem[\protect\citename{Cheng \bgroup et al.\egroup }2016]{self_attention}
Jianpeng Cheng, Li~Dong, and Mirella Lapata.
\newblock 2016.
\newblock Long short-term memory-networks for machine reading.
\newblock {\em arXiv preprint arXiv:1601.06733}.

\bibitem[\protect\citename{Cho \bgroup et al.\egroup }2014]{gru}
Kyunghyun Cho, Bart Van~Merri{\"e}nboer, Dzmitry Bahdanau, and Yoshua Bengio.
\newblock 2014.
\newblock On the properties of neural machine translation: Encoder-decoder
  approaches.
\newblock {\em arXiv preprint arXiv:1409.1259}.

\bibitem[\protect\citename{Devlin \bgroup et al.\egroup }2018]{bert}
Jacob Devlin, Ming-Wei Chang, Kenton Lee, and Kristina Toutanova.
\newblock 2018.
\newblock Bert: Pre-training of deep bidirectional transformers for language
  understanding.
\newblock {\em arXiv preprint arXiv:1810.04805}.

\bibitem[\protect\citename{Hochreiter and Schmidhuber}1997]{lstm}
Sepp Hochreiter and J{\"u}rgen Schmidhuber.
\newblock 1997.
\newblock Long short-term memory.
\newblock {\em Neural computation}, 9(8):1735--1780.

\bibitem[\protect\citename{Kingma and Ba}2014]{adam}
Diederik~P Kingma and Jimmy Ba.
\newblock 2014.
\newblock Adam: A method for stochastic optimization.
\newblock {\em arXiv preprint arXiv:1412.6980}.

\bibitem[\protect\citename{Lample \bgroup et al.\egroup }2016]{char_emb}
Guillaume Lample, Miguel Ballesteros, Sandeep Subramanian, Kazuya Kawakami, and
  Chris Dyer.
\newblock 2016.
\newblock Neural architectures for named entity recognition.
\newblock {\em arXiv preprint arXiv:1603.01360}.

\bibitem[\protect\citename{Liu \bgroup et al.\egroup }2018]{language_model}
Liyuan Liu, Jingbo Shang, Xiang Ren, Frank~Fangzheng Xu, Huan Gui, Jian Peng,
  and Jiawei Han.
\newblock 2018.
\newblock Empower sequence labeling with task-aware neural language model.
\newblock In {\em Thirty-Second AAAI Conference on Artificial Intelligence}.

\bibitem[\protect\citename{Liu \bgroup et al.\egroup }2019]{roberta}
Yinhan Liu, Myle Ott, Naman Goyal, Jingfei Du, Mandar Joshi, Danqi Chen, Omer
  Levy, Mike Lewis, Luke Zettlemoyer, and Veselin Stoyanov.
\newblock 2019.
\newblock Roberta: A robustly optimized bert pretraining approach.
\newblock {\em arXiv preprint arXiv:1907.11692}.

\bibitem[\protect\citename{Pennington \bgroup et al.\egroup }2014]{glove}
Jeffrey Pennington, Richard Socher, and Christopher~D Manning.
\newblock 2014.
\newblock Glove: Global vectors for word representation.
\newblock In {\em Proceedings of the 2014 conference on empirical methods in
  natural language processing (EMNLP)}, pages 1532--1543.

\bibitem[\protect\citename{Peters \bgroup et al.\egroup }2018]{ELMo}
Matthew~E Peters, Mark Neumann, Mohit Iyyer, Matt Gardner, Christopher Clark,
  Kenton Lee, and Luke Zettlemoyer.
\newblock 2018.
\newblock Deep contextualized word representations.
\newblock {\em arXiv preprint arXiv:1802.05365}.

\bibitem[\protect\citename{Radford \bgroup et al.\egroup }2015]{gan}
Alec Radford, Luke Metz, and Soumith Chintala.
\newblock 2015.
\newblock Unsupervised representation learning with deep convolutional
  generative adversarial networks.
\newblock {\em arXiv preprint arXiv:1511.06434}.

\bibitem[\protect\citename{Radford \bgroup et al.\egroup }2019]{gpt2}
Alec Radford, Jeffrey Wu, Rewon Child, David Luan, Dario Amodei, and Ilya
  Sutskever.
\newblock 2019.
\newblock Language models are unsupervised multitask learners.
\newblock {\em OpenAI Blog}, 1(8):9.

\bibitem[\protect\citename{Shirani \bgroup et al.\egroup }2019]{main}
Amirreza Shirani, Franck Dernoncourt, Paul Asente, Nedim Lipka, Seokhwan Kim,
  Jose Echevarria, and Thamar Solorio.
\newblock 2019.
\newblock Learning emphasis selection for written text in visual media from
  crowd-sourced label distributions.
\newblock In {\em Proceedings of the 57th Annual Meeting of the Association for
  Computational Linguistics}, pages 1167--1172.

\bibitem[\protect\citename{Shirani \bgroup et al.\egroup
  }2020]{shirani2020semeval}
Amirreza Shirani, Franck Dernoncourt, Nedim Lipka, Paul Asente, Jose
  Echevarria, and Thamar Solorio.
\newblock 2020.
\newblock Semeval-2020 task 10: Emphasis selection for written text in visual
  media.
\newblock In {\em Proceedings of the 14th International Workshop on Semantic
  Evaluation}.

\bibitem[\protect\citename{Vaswani \bgroup et al.\egroup }2017]{transformer}
Ashish Vaswani, Noam Shazeer, Niki Parmar, Jakob Uszkoreit, Llion Jones,
  Aidan~N Gomez, {\L}ukasz Kaiser, and Illia Polosukhin.
\newblock 2017.
\newblock Attention is all you need.
\newblock In {\em Advances in neural information processing systems}, pages
  5998--6008.

\bibitem[\protect\citename{Yang \bgroup et al.\egroup }2016]{paper2}
Zhilin Yang, Ruslan Salakhutdinov, and William Cohen.
\newblock 2016.
\newblock Multi-task cross-lingual sequence tagging from scratch.
\newblock {\em arXiv preprint arXiv:1603.06270}.

\bibitem[\protect\citename{Yang \bgroup et al.\egroup }2019]{xlnet}
Zhilin Yang, Zihang Dai, Yiming Yang, Jaime Carbonell, Russ~R Salakhutdinov,
  and Quoc~V Le.
\newblock 2019.
\newblock Xlnet: Generalized autoregressive pretraining for language
  understanding.
\newblock In {\em Advances in neural information processing systems}, pages
  5754--5764.

\end{thebibliography}

\end{document}